\documentclass[sigconf]{acmart}

\usepackage{graphicx}
\usepackage{multirow}
\usepackage{subcaption}
\usepackage{tikz}
\usepackage{comment}
\usepackage{color, colortbl}
\usepackage{wrapfig}

\usepackage{amsmath,amsfonts,bm}
\usepackage{acronym}
\usepackage{nicefrac}
\usepackage{xspace}

\acrodef{INN}{Invertible Neural Network}
\acrodef{MLE}{Maximum Likelihood Estimation}
\acrodef{PCC}{Polynomial Color Correction}
\acrodef{GAN}{Generative Adversarial Network}
\acrodef{NLL}{negative log-likelihood}
\acrodef{RL}{Reinforcement Learning}
\acrodef{MLP}{multi-layer perceptron}
\acrodef{SDR}{Standard Dynamic Range}
\acrodef{HDR}{High Dynamic Range}
\acrodef{SM}{supplementary material}
\acrodef{PCA}{Principal Component Analysis}
\acrodef{VAE}{Variational Autoencoder}
\acrodef{CNNs}{Convolutional Neural Networks}

\newcommand{\FiveK}{{MIT-Adobe FiveK}\xspace}
\newcommand{\CIEDE}{{CIE\,DE\,2000}\xspace}

\newcommand{\figref}[1]{Fig.~\ref{fig:#1}}
\newcommand{\twofigref}[2]{Fig.~\ref{fig:#1} and~\ref{fig:#2}}
\newcommand{\secref}[1]{Sec.~\ref{sec:#1}}

\newcommand{\tableref}[1]{Table~\ref{tab:#1}}

\newcommand{\equationref}[1]{Eq.~\ref{eq:#1}}

\newcommand{\rev}[1]{#1}

\usepackage{breakcites}

\usepackage{arydshln}
\usepackage{tabularx}
\newcolumntype{Y}{>{\centering\arraybackslash}X}

\AtBeginDocument{%
  \providecommand\BibTeX{{%
    \normalfont B\kern-0.5em{\scshape i\kern-0.25em b}\kern-0.8em\TeX}}}

\copyrightyear{2022}
\acmYear{2022}
\setcopyright{rightsretained}
\acmConference[CVMP '22]{European Conference on Visual Media Production}{December 1--2, 2022}{London, United Kingdom}
\acmBooktitle{European Conference on Visual Media Production (CVMP '22), December 1--2, 2022, London, United Kingdom}\acmDOI{10.1145/3565516.3565520}
\acmISBN{978-1-4503-9939-5/22/12}
\begin{document}

\title{Distilling Style from Image Pairs for Global Forward and Inverse Tone Mapping}


\author{Aamir Mustafa}
\affiliation{%
  \institution{University of Cambridge}
  \country{UK}
}
\email{aamir.mustafa@cl.cam.ac.uk}

\author{Param Hanji}
\affiliation{%
  \institution{University of Cambridge}
  \country{UK}
}
\email{param.hanji@cl.cam.ac.uk}

\author{Rafa{\l} K. Mantiuk}
\affiliation{%
  \institution{University of Cambridge}
  \country{UK}
}
\email{rafal.mantiuk@cl.cam.ac.uk}

\renewcommand{\shortauthors}{Mustafa et al.}

\begin{abstract}
Many image enhancement or editing operations, such as forward and inverse tone mapping or color grading, do not have a unique solution, but instead a range of solutions, each representing a different style. Despite this, existing learning-based methods attempt to learn a unique mapping, disregarding this style. In this work, we show that information about the style can be distilled from collections of image pairs and encoded into a 2- or 3-dimensional vector. This gives us not only an efficient representation but also an interpretable latent space for editing the image style. We represent the global color mapping between a pair of images as a custom normalizing flow, conditioned on a polynomial basis of the pixel color. We show that such a network is more effective than PCA or VAE at encoding image style in low-dimensional space and lets us obtain an accuracy close to 40\,dB, which is about 7-10\,dB improvement over the state-of-the-art methods.
\end{abstract}




\begin{teaserfigure}
  \includegraphics[trim={0cm 3.3cm 1.7cm 3.5cm}, clip, width=0.95\linewidth]{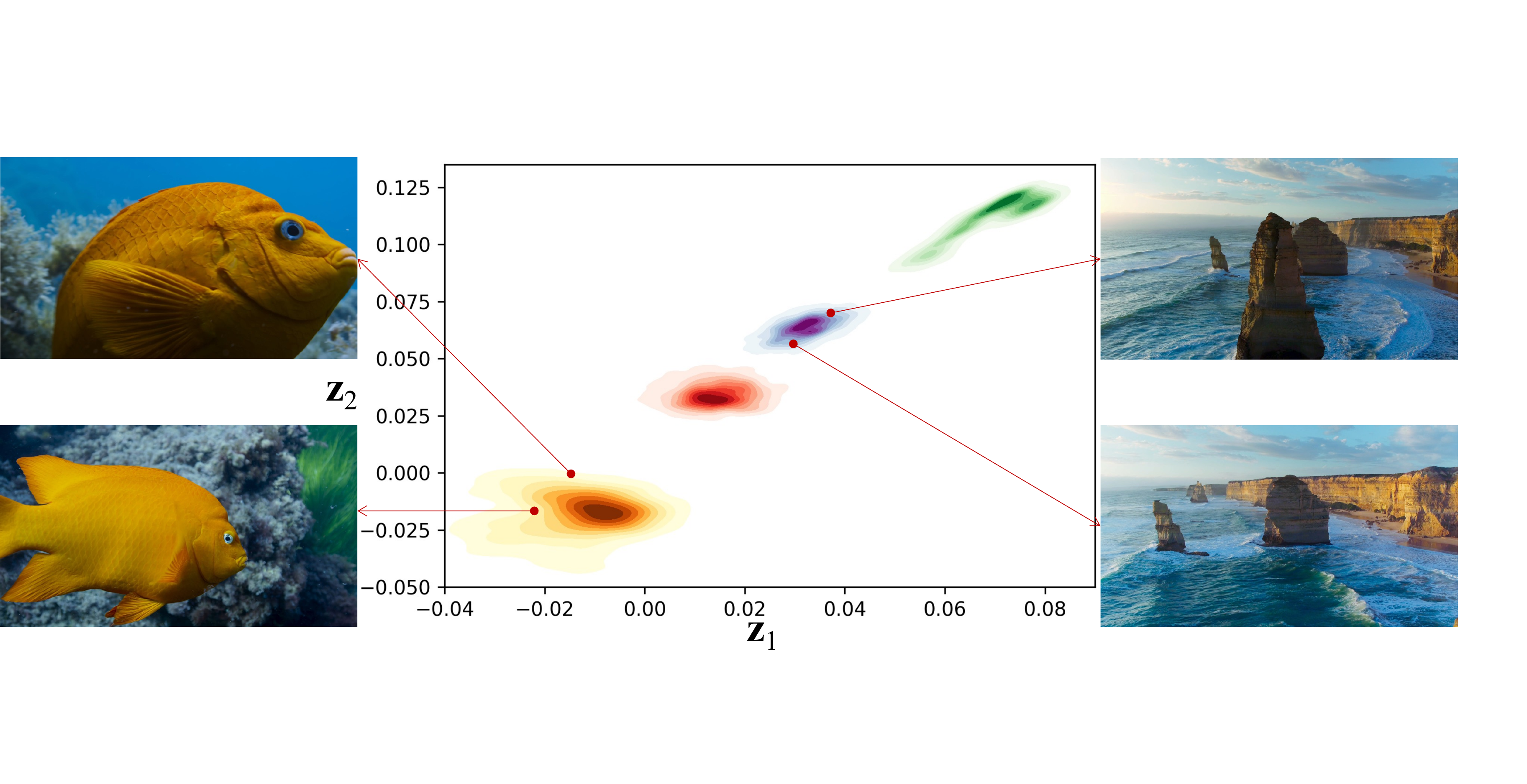}
  \caption{Distribution of the latent color code representation for different frames. We see that frames with similar content and similar color retouching have been encoded close together in the latent space. All frames are taken from the BBC documentary Blue Planet II Episode ``Green Seas". Note that the model trained to encode the style of frames in 2 dimensions (see \secref{dim-style-code}) is used in this plot for better visualization. }
  \vspace{1em}
  \label{fig:teaser}
\end{teaserfigure}

\maketitle

\section{Introduction}


Several applications in image and video processing require mapping source to target colors, such as mapping \ac{HDR} to \ac{SDR} frames (tone mapping), \ac{SDR} to \ac{HDR} frames (inverse tone mapping), RAW to color-graded frames or retouched images. Tone mapping and inverse tone mapping have become particularly relevant with the introduction of \ac{HDR} formats and standards, which necessities mastering content separately for both \ac{SDR} and \ac{HDR} devices for best presentation.

The research on forward and inverse tone mapping dates several decades \cite{StockhamJr1972,Tumblin1993,Reinhard2002,Mantiuk2008,Eilertsen2017}. Most tone mapping methods propose hand-crafted recipes \cite{StockhamJr1972,Tumblin1993,Reinhard2002}, or optimization criteria \cite{Mantiuk2008}, which should produce the most desirable image. Rather than following this line of research to find the right recipe, we learn the mapping from collections of (source, target) image pairs manually prepared by skilled color artists. With the advent of deep learning and large datasets, recent works treat the relationship between paired images as a one-to-one mapping, which they typically learn with \ac{CNNs}.
The notable conceptual flaw of this approach is that tone mapping, color grading, retouching, and inverse tone mapping do not have a single solution. If the task is given to several skilled color artists, each one is likely to produce a different result. This was shown both in early research on tone mapping \cite{Yoshida2006}, and in the Adobe-MIT 5k dataset \cite{Bychkovsky2011}, where each image was differently retouched by 5 photographers. Each color artist conveys a different style, and methods that automate this process should respect and preserve such intended style. \rev{Because the style information is not present in the input image, existing methods that learn a CNN based image-to-image mapping can learn at most a single style. The possibility of an indefinite number of equally plausible solutions, makes a CNN based image-to-image translation network incapable of modelling the intended style without an additional meta-data as style encoding.
}

\rev{
To this end, in this work, we deviate from the conventional approach of learning an one-to-one mapping function from the source to the target domain. Instead, we propose an orthogonal direction of research to distill the artistic style from pairs of source and target images and encode it in a low dimensional \emph{style} vector. More specifically, we present a customized conditional \ac{INN}, which captures the one-to-many nature of the problem. Unlike other works, our lightweight architecture operates in a pixel-wise manner to map the target pixels to a latent style, given the source pixels as conditioning inputs. The proposed bi-directional training, lets us converge the pixel-wise mapping to a single low dimensional latent \emph{style} vector as the meta data for the entire image. As shown in~\figref{teaser}, our trained \ac{INN} can map images of similar styles to neighboring points in the latent style space. The proposed method allows us to distill the underlying global mapping between the source and the target image pair into a latent representation vector at \emph{inference} time. 
}

The main benefit of extracting style is that it lets us efficiently encode very precise color mapping from a source to target image. Such an encoding could be used for simultaneous transmission of HDR and SDR content, where each \ac{HDR} frame is accompanied by a 2--4 dimensional vector, which lets us reconstruct \ac{SDR} frames with almost perfect accuracy ($\sim$40\,dB). The encoding can also be used to easily edit the color mapping (color grading, retouching, tone-mapping) using just a pair of sliders instead of much more complex interfaces used for those tasks. 
Finally, if we wish to perform fully automated mapping in an ``average" style, we can run inference on our network while setting the style vector to 0. \\

The main contributions of this work are:
\begin{itemize}
    \item We show that it is necessary to distill a style from image pairs to effectively learn a highly accurate color mapping ($\sim$40\,dB) and that the style can be encoded in a 2--4 dimensional vector. 
    
    \item We propose an \ac{INN}-based generative model that can learn the distribution of styles and encode it in a low-dimensional vector. The model can be trained on moderately sized datasets (500 image pairs) and achieves a dramatic gain in performance of over 10\,dB compared to the methods that disregard style.

    
    \item The utility of the color mapping with a style vector is demonstrated in video transmission, semi-automatic color-grading, and inverse tone mapping. 
\end{itemize}

\rev{In this work, we distill the style between the source-target image pairs, which have been manually color graded by color artists. Such color mappings for video content are only global in nature (the same for all pixels in an image, $\mathcal{M}: \mathcal{R}^3 \rightarrow \mathcal{R}^3$) and do not contain any local (spatially varying) changes.  However, our method can be easily extended to a manual local mapping, by splitting frames into a number of tiles, as done in \cite{Eilertsen2015}.}  



\section{Related work}
\label{sec:related_work}

Our work addresses the problem of learning color mapping from pairs of images, found in tone-mapping \cite{Bychkovsky2011,gharbi2017deep,Rana2020}, inverse tone mapping \cite{Eilertsen2017a,Marnerides2018expandnet,Liu2020,Santos2020}, image enhancement and automatic retouching \cite{Yan2016,park2018distort,wang2019underexposed,zeng2020learning,he2020conditional,kim2021representative}. The majority of existing methods attempt to learn one-to-one mapping from a large collection of image pairs, such as that found in \FiveK dataset \cite{Bychkovsky2011}. The mapping can be represented using polynomial basis functions \cite{Yan2016}, 3D LUTs with trainable weights \cite{zeng2020learning}, a linear combination of representative colors \cite{kim2021representative}, color-to-color \ac{MLP} \cite{he2020conditional}, an encoder-decoder architecture \cite{chen2018deep, ignatov2017dslr, Yan2016, wang2019underexposed}, a bilateral grid of affine color transformation matrices \cite{gharbi2017deep}, or a \ac{RL} policy that mimics the sequence of operations that a human expert would take \cite{park2018distort}. All those methods attempt to extract local and global features from an input image so that the mapping function can adapt to image content. We show that this information alone is insufficient to obtain a highly accurate mapping.

Several works address the problem of learning different styles of image mapping, but none attempt to achieve the same goals as our work. PieNet \cite{kim2020pienet} and StarEnhancer \cite{song2021starenhancer} learn embedding of a style (or personalization) of image-to-image mapping. They assume that all images processed by a single expert (from the Adobe-MIT-5K dataset) share the same style. We found this assumption to be overly optimistic, as experts often vary style between images. For that reason, we assign a separate style vector to each image pair. This lets us obtain an accuracy close to 40\,dB, while both methods report only about 25\,dB on the same dataset. Neither of the methods attempts to reduce the dimensionality of the style vector to make it better suited to coding and manipulation of images. 


\section{Methodology}
\label{sec:approach}


\subsection{Problem formulation}
\label{sec:problem-formulation}
We aim to model a global, spatially invariant color mapping between pairs of images or video frames. Furthermore, such a mapping should be controlled by a low-dimensional \emph{style} vector. 
Let $\{\mathbf{x}^i, \mathbf{y}^i\}_{i=1}^N$ represent $N$ images pairs. Each RGB pixel $p$ in those pairs is related by a global mapping $\mathcal{M}$:
\begin{equation} \label{eq:image-mapping}
    \mathbf{x}^i_p = \mathcal{M}(\mathbf{y}^i_p, \mathbf{z}^i) \,,
\end{equation}
where $i$ is the image index and $\mathbf{z}^i$ is the low-dimensional vector that encodes \emph{style} parameters specific to each pair. Please note that we use $\mathbf{y}$ for source and $\mathbf{x}$ for target color to be consistent with the normalizing flow notation used later. 

Such a mapping can be effectively expressed as a parametric function using \ac{PCC} basis functions~\cite{finlayson1997constrained, finlayson2015color}: 
\begin{equation} \label{eq:style-matrix}
    \mathbf{x}^i_p = \mathcal{C}(\mathbf{y}^i_p) \cdot \mathbf{M}^i \,,
\end{equation}
where $\mathcal{C}(\mathbf{y}^i_p)$ is the polynomial basis function for the pixel $\mathbf{y}^i_p$:
\begin{equation} \label{eq:condition}
    \mathcal{C}(\mathbf{y}_p^i) = \mathcal{C}([r,g,b]) = [r, g, b, r^2, g^2, b^2, rg, gb, br, \ldots , r^4, g^4, b^4]\,,
\end{equation}
and $\mathbf{M}^i$ is the $34{\times}3$ style matrix (a representation of $\mathbf{z}^i$). We found that at least a 4\textsuperscript{th} degree \ac{PCC} with $34 \times 3 = 102$ parameters is needed for our diverse set of scenes (further discussed in \secref{ablation-conditioning}). However, such a large latent representation is not interpretable and cannot be intuitively modified to edit the style. While the obvious approach is to use one of the  dimensionality reduction techniques such as \ac{PCA} and \ac{VAE}, we found them to be ineffective for our problem, as we will show in \secref{pca-vs-autoencoder}. Instead, we propose to use a custom adaptation of normalizing flows~\cite{rezende2015variational} to represent the style mapping.
\subsection{Our approach}
We cast the problem of finding a suitable approximation of $\mathcal{M}$ from~\equationref{image-mapping} as a generative modeling task using supervised learning. We learn a conditional mapping from a known latent distribution to the target domain using a parametric model $g_\theta$ with weights $\theta$. After training, the learned network $g_\theta$ and extracted latent vector $\mathbf{z}^i$ reproduce the required per-pixel mapping,
\begin{equation} \label{eq:pixel-mapping}
    \mathbf{x}^i_p = g_\theta(\mathbf{z}^i; \mathbf{c}^i_p ) \,.
\end{equation}
The conditioning vector $\mathbf{c}^i_p$ could be equal to the source pixel value, $\mathbf{c}^i_p=\mathbf{y}^i_p$. However, we found that performance is much better if the conditioning vector contains the polynomial basis functions:
$\mathbf{c}^i_p=\mathcal{C}(\mathbf{y}^i_p)$.
Intuitively, when the polynomial basis is supplied, the network only needs to learn the appropriate linear combination, similar to matrix $\mathbf{M}^i$ from \equationref{style-matrix}. This dimensionality expansion approach is akin to positional encoding, used to represent multi-dimensional functions in NeRF methods \cite{Mildenhall2020,tancik2020fourier}. We also observed that adding global information to the conditioning vector, such as the image histogram or VGG features~\cite{simonyan2014vgg} to $\mathbf{c}^i_p$ has little effect, as shown in the ablation studies in~\secref{ablation-conditioning}.

\subsection{The latent distribution model} \label{sec:generative-model}
To approximate the global mapping $\mathcal{M}$ with a parametric per-pixel function. We choose the class of generative models called normalizing flows~\cite{rezende2015variational,dinh2016density,kingma2018glow}. They employ invertible layers with tractable determinants of jacobian to learn a non-trivial mapping from a known latent to the target distribution. The bidirectionality of normalizing flows makes them particularly suitable since we wish to both (1) extract a latent \emph{style} vector from a set of pixels (reverse discriminative pass) and (2) reconstruct a pixel from the given input and latent vector (forward generative pass). This enables us to manipulate the image style by converting it to a latent space as an intermediate step (see~\figref{interpretable-latent}).

Given many samples from the target distribution $X$, the objective is to train $g_\theta$ using \ac{MLE} to convert them into samples of a latent distribution $Z$. If $g_\theta$ is invertible, the change of variables formula expresses the probability of a target sample $\mathbf{x} \sim X$ as a function of the probability of the transformed latent $\mathbf{z} \sim Z$:
\begin{equation} \label{eq:change-of-variables}
    p_X(\mathbf{x|\mathbf{c}}) = p_Z\left(g_\theta^{-1}(\mathbf{x};\mathbf{c})\right) \cdot \left|\textrm{det} \nabla g_\theta^{-1}(\mathbf{x};\mathbf{c}) \right|\,.
\end{equation}
Here $\mathbf{c}$ is a conditioning vector, which corresponds to the polynomial coefficients $\mathbf{M}^i$ in our implementation (see \equationref{style-matrix}).

If the latents follow a known distribution such as a standard normal or uniform, \equationref{change-of-variables} gives us an exact expression for the probability of any data point. For efficient \ac{MLE} training, $g_\theta$ should be differentiable, and the determinant of its jacobian should be easy to compute. In practice, a normalizing flow is a deep architecture consisting of invertible layers whose jacobians are diagonal, lower-triangular, or the identity matrix.

\subsection{Network architecture} \label{sec:network}

\begin{figure*}[t]
    \centering
    \includegraphics[width=\linewidth]{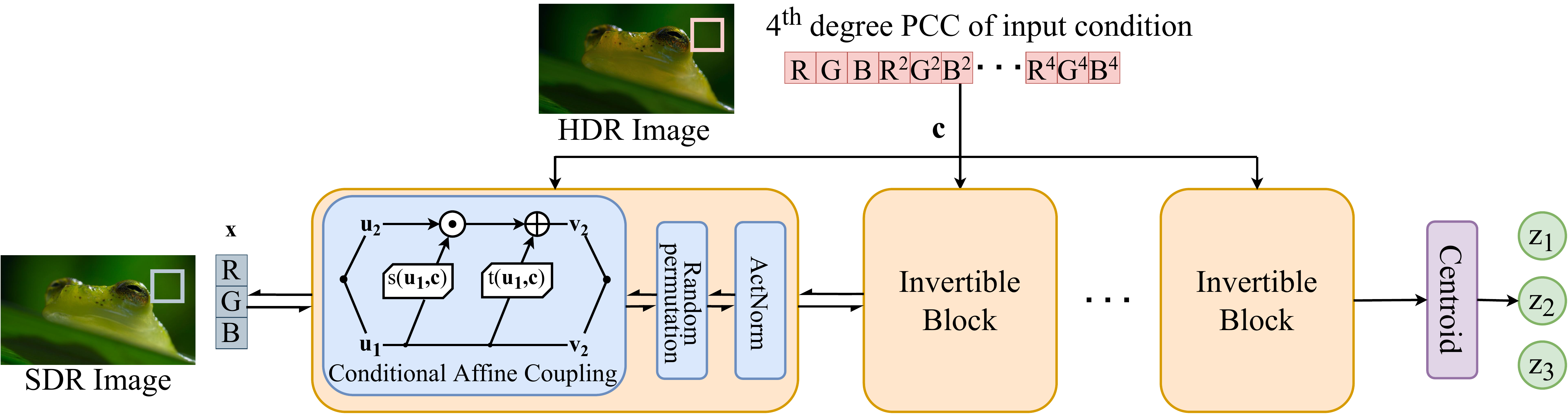}
    \caption{ {The network architecture used for our pixel-wise conditional \ac{INN}. Here the sub-networks $s(\cdot)$ and $t(\cdot)$ are fully connected networks with 2 hidden layers each. We use 8 invertible blocks in our architecture.}}\label{fig:architecture}
\end{figure*}

Our \ac{INN}, depicted in \figref{architecture}, is composed of a series of 8 invertible blocks each consisting of (1) affine coupling layer, (2) random permutation and (3) batch normalization with ActNorm \cite{kingma2018glow}. The coupling layers \cite{dinh2016density} increase the expressive power of the \ac{INN} by incorporating complex, non-invertible sub-networks $s(\cdot)$ and $t(\cdot)$. To keep the total number of trainable parameters small, we use simple 2-layer \ac{MLP}s for both $s(\cdot)$ and $t(\cdot)$.

\subsubsection{Conditional coupling}
Affine coupling~\cite{dinh2016density} is a key operation of most normalizing flow architectures that increases the expressibility of the network for unconditional generative modeling. This involves splitting the inputs $\mathbf{u}$ of an intermediate layer into two equal parts $\mathbf{u}_1$ and $\mathbf{u}_2$. The first, $\mathbf{u}_1$ is transmitted unchanged, while $\mathbf{u}_2$ goes through an affine transformation where parameters are some functions of $\mathbf{u}_1$, realized through complex, non-linear subnetworks $s(\mathbf{u}_1)$ and $t(\mathbf{u}_1)$.

However, we are interested in modeling the conditional distribution $p_X(\mathbf{x}|\mathbf{c})$ as described in \equationref{change-of-variables}. For this, we use conditional coupling described in~\cite{ardizzone2020conditional,lugmayr2020srflow} by appending $\mathbf{c}$ from~\equationref{condition} to the inputs of $s(\cdot)$ and $t(\cdot)$. Now, the output of the conditional coupling layer is given by the concatenated vector [$\mathbf{v_1}$, $\mathbf{v_2}$] where

\begin{equation}
\label{eq:coupling}
    \mathbf{v_1} = \mathbf{u_1}\,, \qquad
    \mathbf{v_2} = \mathbf{u_2} \odot \textrm{exp}({s([\mathbf{u_1}, \mathbf{c}])) + t([\mathbf{u_1}, \mathbf{c}])}\,.
\end{equation}

Since the subnetworks in a coupling block are never inverted themselves, we can append $\mathbf{c}$ without losing the invertibility of the \ac{INN}.
Moreover, the operation has a lower-triangular jacobian whose determinant is the product of diagonal elements.

\subsubsection{Dimensionality of latent}
\label{sec:dim-style-code}

\begin{figure*}[t]
    \centering
    \includegraphics[width=\linewidth]{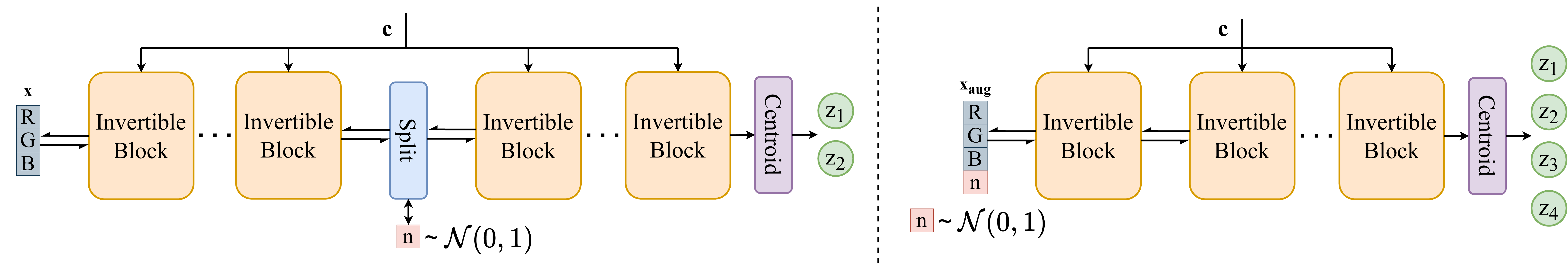}
    \caption{ {Changes in INN architecture to decrease the latent style vector to 2 dimensions (left) or increase to 4 dimensions (right).}}
    \label{fig:dimension-change}
\end{figure*}

Central to an efficient style mapping framework is the dimensionality of the latent vector. Mapping the style of an image from the target domain to a low-dimensional \emph{style} vector allows easy user-interactive image manipulation and editing. Our default architecture depicted in~\figref{architecture} encodes style in 3 latent dimensions, matching the dimensionality of an input pixel. However, some applications may require differently sized \emph{style} vectors. As alternatives, we demonstrate the adaptations that enable the \ac{INN} to operate with fewer or more latent dimensions.

To encode style in fewer dimensions, we split off some features after 4 invertible blocks. As depicted in \figref{dimension-change} (left), these features are forced to follow the standard normal distribution using \ac{MLE}. The remaining features continue through 4 more invertible blocks. The total number of invertible blocks is 8, to match our 3-dimensional INN from \figref{architecture}. This splitting of intermediate features is similar to the multiscale version of the normalizing flow described in~\cite{dinh2016density}.

For a higher-dimensional latent style vector, we construct an invertible model on an augmented input space $\mathbf{x}_\textnormal{aug}$. Similar to~\cite{huang2020augmented}, each input vector $\mathbf{x}$ is appended with samples from a standard normal distribution as shown in~\figref{dimension-change} (right). This allows us to improve the expressibility of the latent space at the cost of an increased dimension (from 3 to 4) for final image manipulation. 

\subsection{Optimization and inference}
\label{sec:optimization-inference}


Similar to other normalizing flows, we train our \ac{INN} by \ac{MLE} where the likelihood is given by~\equationref{change-of-variables}. The network thus learns to conditionally map a target pixel to a latent vector. However, when presented with an entire frame, there is no easy way to extract a single low-dimensional vector that captures style. For that, we augment \ac{MLE} training with a reconstruction loss to force the per-pixel representations of the same frame to lie closer in the latent space (see \figref{teaser}). 

\textbf{Likelihood loss:}
To restrict the magnitude of gradients for backpropagation, it is customary to minimize the \ac{NLL} instead of likelihood since the logarithmic transform is monotonic.
The \ac{NLL} loss is:
\begin{equation}
\begin{split}
    \mathcal{L}_\textrm{NLL}(\mathbf{x}_p,\mathbf{y}_p) &= -\log{p_Z \big(g_\theta^{-1}(\mathbf{x}_p; \mathbf{c}_p)\big)} - \log{\left|\nabla  g_\theta^{-1}(\mathbf{x}_p; \mathbf{c}_p) \right|} \\
    &= \frac{\log{2\pi} + \big( g_\theta^{-1}(\mathbf{x}_p; \mathbf{c}_p) \big)^2}{2} - \log{\left|\nabla  g_\theta^{-1}(\mathbf{x}_p; \mathbf{c}_p) \right|}
\end{split}
\end{equation}
This result follows since we choose a standard normal $\mathbf{z}_p{\sim}\mathcal{N}(0,1)$ as the base distribution with the following log-likelihood:
\begin{equation}
    \log{p_Z(z) = -\frac{1}{2}~\log{2\pi} - \frac{1}{2} ~||~z~ ||^2_2 }.
\end{equation}

\begin{figure*}[h]
    \centering
  {\includegraphics[trim={1.25cm 8.2cm 8cm 2.70cm}, clip, width=0.95\linewidth]{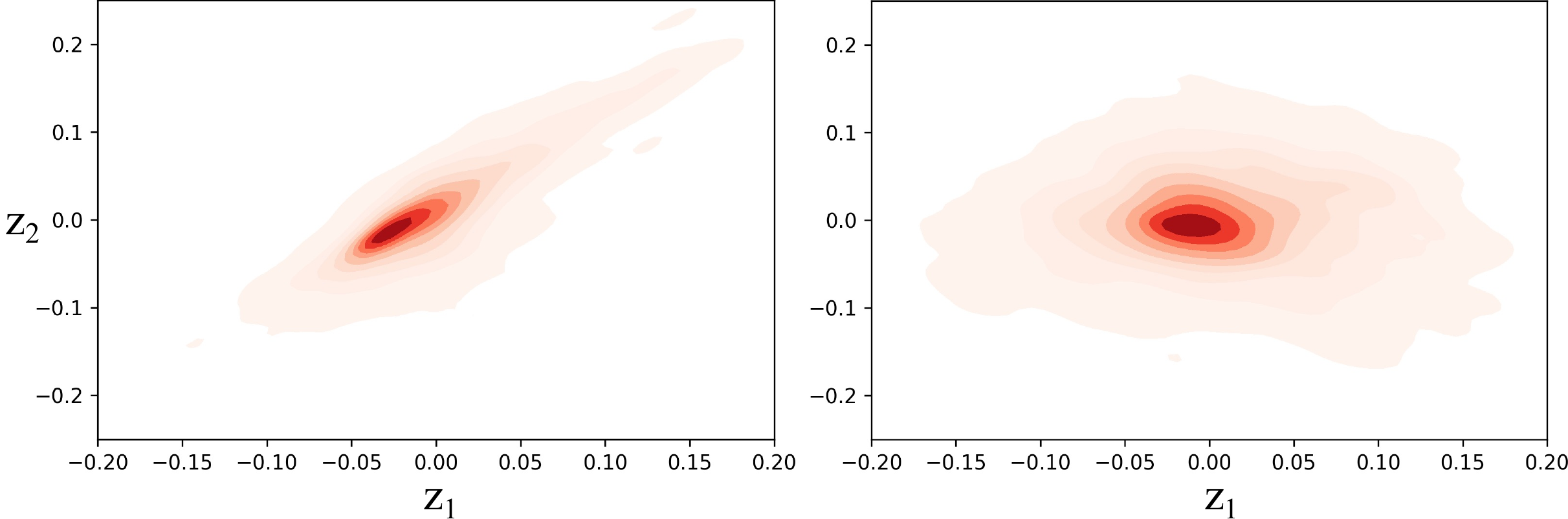} }
    \caption{ {The figure shows the distribution of the latent color representation for different frames from the training set with (left) and without (right) the reconstruction loss $\mathcal{L}_\textrm{rec}$. We see that the latent vectors of the training set follow a normal distribution. Frames are taken from the BBC documentary Planet Earth ``Jungles" and the INN trained to encode the style of frames to 2 dimensions is used for better visualization.}} \label{fig:training_plot}
        
\end{figure*}

The first term is a constant w.r.t. z and can be dropped during training. Due to this loss, the \ac{INN} learns a bijective mapping from the distribution of pixels to the chosen latent, conditioned on the encoded input pixel. Through \ac{MLE}, we encourage per-pixel latents to follow a standard normal distribution as shown in~\figref{training_plot} (right). 

\textbf{Reconstruction loss:}
Our secondary requirement is for pixels of the same frame to cluster together in the latent space $Z$.
We achieve this by first passing all $K$ pixels of frame $i$ through the INN in reverse, and computing the centroid of their latent representations,
\begin{equation} \label{eq:single-latent}
    \mathbf{z}^i = \frac{1}{K}\sum_{p=1}^{K} g_\theta^{-1}(\mathbf{x}_p^i; \mathcal{C}(\mathbf{y}^i_p))\,,
\end{equation}
where $\mathbf{z}^i$ is the per-frame \emph{style} vector. Then, we reconstruct the frame with $K$ forward passes of the INN using the single extracted \emph{style} latent $\mathbf{z}^i$ but different conditional inputs. Finally, we compute the reconstruction loss $\mathcal{L}_\textrm{rec}$ for each pixel of an image as follows:

\begin{equation}
    \mathcal{L}_\textrm{rec}(\mathbf{x}^i_p,\mathbf{y}^i_p) = \parallel g_\theta(\mathbf{z}^i; \mathcal{C}(\mathbf{y}^i_p)) - \mathbf{x}^i_p \parallel_2
\end{equation}
The reconstruction loss enforces a similar value of the style vector $\mathbf{z}^i$ for the entire frame.
We also observe that this simple constraint on the latent representation allows mapping of dissimilar frames in the input space to distant and distinct regions in the latent space (see \figref{teaser}).



Optimization of our INN is done bi-directionally with the total loss as the sum of the \ac{NLL} loss $\mathcal{L}_\textrm{NLL}$ and the reconstruction loss $\mathcal{L}_\textrm{rec}$.
After successful training, the INN can be used to extract an overall per-frame \emph{style} vector. This is achieved by running a reverse pass for each pixel of a given image and computing the centroid according to \equationref{single-latent}. In \figref{training_plot}, we show the change in the latent space for the training samples due to the addition of $\mathcal{L}_\textrm{rec}$.

\section{Results}
\label{sec:results}

In this section, we evaluate the efficacy of our method in encoding the style of a target domain image into a low-dimensional latent space for the task of forward and inverse tone mapping. 
First, we compare our approach of conditioning color mapping on a style with the traditional approach of conditioning on the input image (\secref{image_content_vs_style}, comparison with HDRNet). 
Then, we demonstrate that the existing combination of \ac{PCC} with dimensionality reduction approaches (PCA and VAE) gives far inferior results as compared to our \ac{INN} (\secref{pca-vs-autoencoder}). 
We report results in terms of PSNR (for RGB values) and FLIP~\cite{andersson2020flip} here, and CIELab in the appendix. We choose these metrics because they are sensitive to color differences (unlike SSIM). 


\begin{figure*}[t]
    \centering
    {\includegraphics[trim={0cm 33.25cm 1.4cm 0cm}, clip, width=0.95\linewidth]{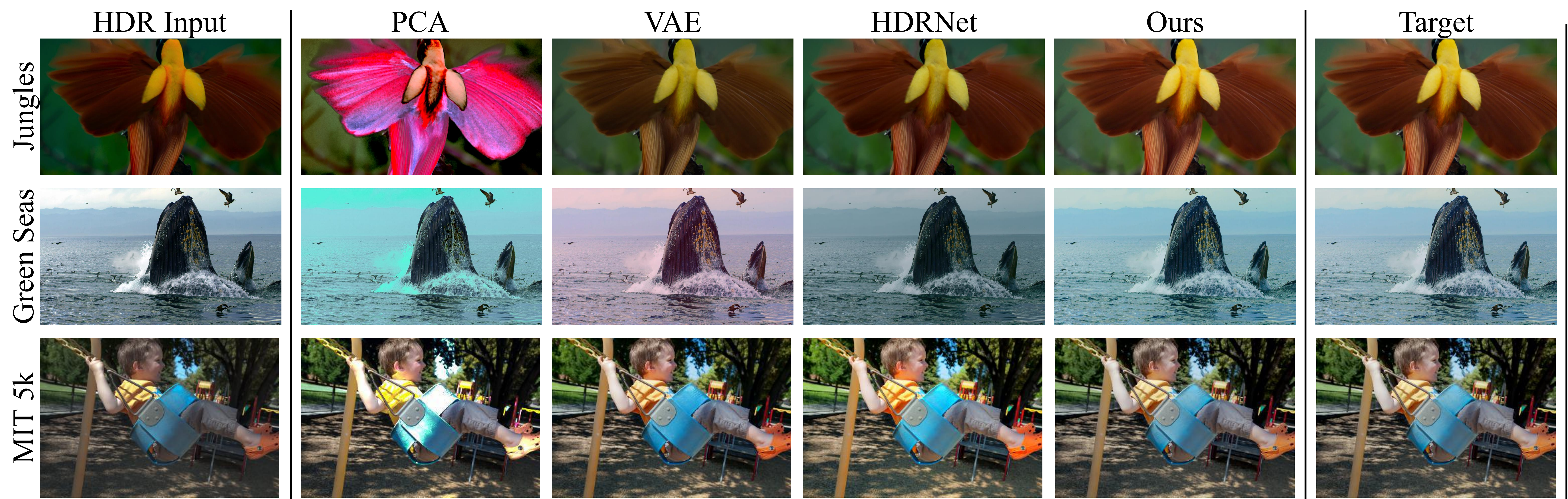} }
    \caption{  {Qualitative comparisons with different methods on 3 datasets for the task of forward tone mapping. The target for MIT5k dataset is the expert retouched image. Additional results are provided in the appendix.}}
    \label{fig:qualitative_results}
    
\end{figure*}

\subsection{Datasets}
\label{sec:data}

We rely on two sources of SDR-HDR pairs. For images, we use Adobe-MIT 5K dataset \cite{Bychkovsky2011}. Each 
RAW image in this dataset was tone-mapped (retouched) by 5 experts, who produced results in different styles. We report the results for expert C in the main paper, and for other expert in the supplementary materials. 
All images are rescaled to the height of $480$ pixels for faster training and then split into a random 80/20\% train/test sets. Due to the pixel-wise formulation of our \ac{INN},
the same trained model can be employed on images of different resolutions at inference time.


Due to the lack of any publicly available manually color graded video datasets, we decoded 3 Blu-ray movies, namely ``BBC Planet Earth II Episode 3 - Jungles", ``BBC Blue Planet II Episode 5 - Green Seas" and ``The Lego Batman Movie". 4K HDR content is often sold with two disks — one color graded for 4K HDR and another for 1080p SDR displays. We took advantage of that by extracting content from both disks. The frames from SDR and HDR streams were time-synchronized by finding the offset that maximized cross-correlation. Finally, the frames were manually inspected to ensure close correspondence.
For good diversity, we construct a sequence from each video by collecting every 120\textsuperscript{th} frame. For each video sequence, the first 80\% of frames are used for training and the remaining 20\% for testing. On average, we have 500-800 frames in the training set per movie.
Each frame is rescaled to a resolution of half HD ($960 \times 540$).
Both SDR and HDR RGB pixel values are display encoded (BT.2020 + PQ for HDR, BT.709 + sRGB for SDR). 

\rev{
Publicly available datasets proposed in methods like \cite{kim2019deep,chen2021new} do not include manual color grading, but instead rely on Youtube's automatic HDR to SDR conversion process. The primary objective of our work is to model manual color grading, making such datasets unsuitable for the task of distilling color artist's style.
}

\subsection{Conditioning on style vs. image content}
\label{sec:image_content_vs_style}

The central assumption of current deep learning tone mapping methods is that the right mapping can be found by analyzing global and local image features \cite{gharbi2017deep}. Based on that assumption, the existing methods employ large convolutional and/or fully connected networks operating on an entire image. In contrast, our INN operates at the pixel level to effectively distill the image-specific style vector without analyzing the image content. 
Here, we test which approach can better predict the results of manual tone-mapping and color grading. We train our INN separately on pairs of HDR-SDR frames taken from each movie from the movies dataset (see \secref{data}). Our pixel-wise training scheme allows us to operate on high-resolution images. Inference for each frame of half HD resolution ($960 \times 540$)
takes 0.025~seconds. Each model is trained for 80 epochs with an initial learning rate of $5e-4$ with gradual learning rate scheduling. Additionally, for a more challenging setting, we train our INN on the Adobe-MIT 5K dataset \cite{Bychkovsky2011}. 

\begin{figure*}[t]
    \centering

    \includegraphics[trim={0cm 46.7cm 22.4cm 0cm}, clip, width=0.995\linewidth]{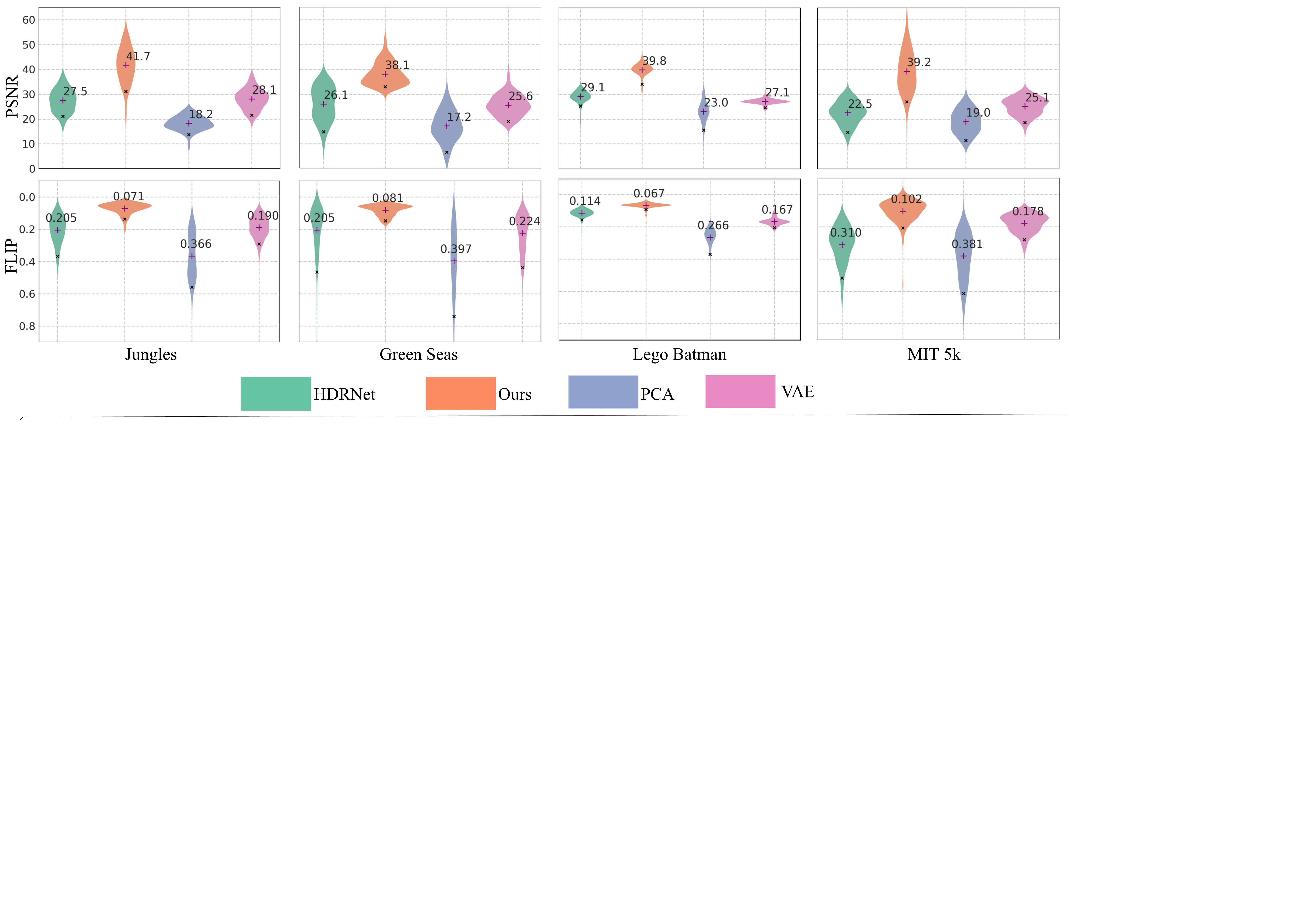}
    \caption{ {Comparison of results on different datasets for the task of forward tone mapping. Note that our method achieves a substantial improvement in performance compared to other dimensionality reduction methods across datasets. The purple `$+$' in the plots show the mean and black `$\times$' show the lowest 5$^{th}$ percentiles. Note that the y-axis for FLIP metric has been reversed. The results for Ours, PCA and VAE are reported for the model that encodes the style representation into 3 dimensional latent vector.
    }}\label{fig:results_tm}
        
    \includegraphics[trim={0cm 66.5cm 46cm 0cm}, clip, width=0.8\linewidth]{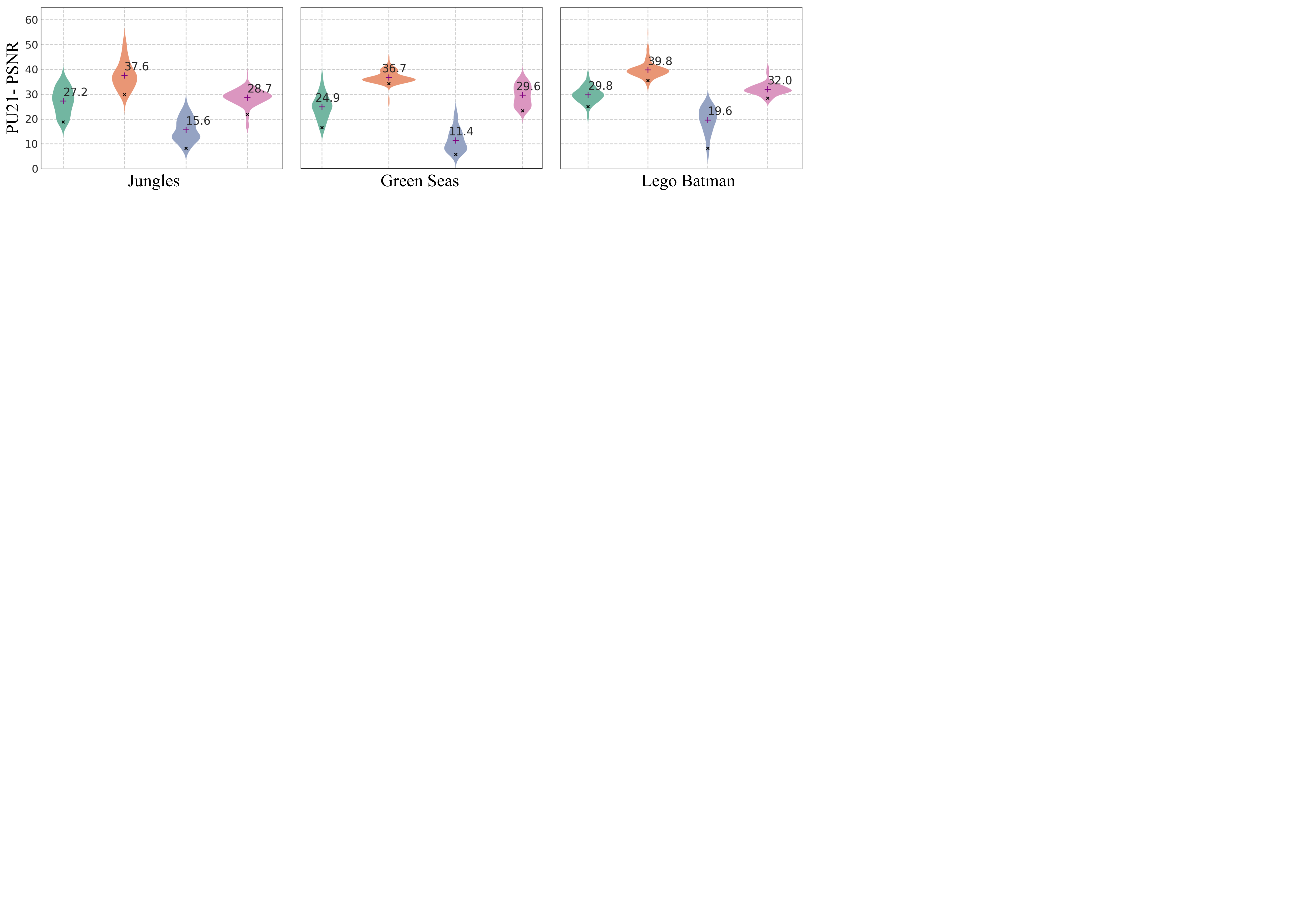}
    \caption{ {Comparison of results on different datasets for the task of inverse tone mapping on the movies dataset. To adapt PSNR to HDR images, we use perceptually uniform PU21 transform~\cite{mantiuk2021pu21}. The labels for the violin plots are consistent with \figref{results_tm}. The results for Ours, PCA and VAE are reported for the model that encodes the style representation into 3 dimensional latent vector.} 
    }
    \label{fig:results_itm}
\end{figure*}

\begin{table*}[t]
\caption{{Comparison of our \ac{INN} with state-of-the-art methods that assume one-to-one mapping for Expert C from the MIT5k dataset~\cite{Bychkovsky2011}. The substantial improvement in performance clearly demonstrates how distilling style can lead to almost perfect reconstruction accuracy. The values for our and HDRNet methods were obtained using the same train/test splits, and the values from other methods are taken from the respective works. 
}}

\label{tab:mit5k_sota}
\centering

\resizebox{.995\textwidth}{!}{
\scriptsize
    \begin{tabularx}{\textwidth}{@{}YYYYYYYYYYY@{}}
    \toprule
    \rowcolor{black!10} & HDRNet & UPE & GleNet & 3DLUT & StarEnh & Curl & DPE & CRSNet & DLPF & Ours \\
    \rowcolor{black!10} & \cite{gharbi2017deep} & \cite{wang2019underexposed} & \cite{kim2020global} & \cite{zeng2020learning} & \cite{song2021starenhancer} & \cite{moran2020curl} & \cite{chen2018deep} & \cite{he2020conditional} & \cite{Moran_2020_dlpf}& \\
    \midrule
    
    PSNR & 22.49 & 23.24 & 25.88 & 24.92  & 25.46 & 24.04 & 23.76 & 24.23 & 23.93 &\textbf{39.22} \\
    Params. & 482K & 1M & - & $<$600K  & 14M & 1.4M & 3.34M & 36K & 1.8M &\textbf{31.4K} \\

    \bottomrule
    \end{tabularx} 

    }
\end{table*}

As a representative example of existing learning-based tone mapping, we compare our results with HDRNet \cite{gharbi2017deep} (implemented in PyTorch \cite{JinchenHDRNet}), retrained on the same data as our method. The numerical results and quality metric distributions, shown in~\figref{results_tm}, demonstrate a dramatic improvement of 10-16\,dB as compared to HDRNet. This shows that information about the style is necessary to faithfully reproduce manually color-graded or retouched images. Although HDRNet is a much larger network, consisting of 482K trainable parameters compared to 31K for our \ac{INN}, it cannot infer the target image based on the source image alone. A few example images shown in \figref{qualitative_results} demonstrate that HDRNet fails to reproduce accurate color and tones of the target images.
Similar to HDRNet, we see from~\tableref{mit5k_sota} that other state-of-the-art image enhancement methods that do not model style struggle to reconstruct image-specific retouching with an accuracy higher than 25\,dB. 
\rev{
Please note that due to the novelty of our approach of encoding style as meta data, a direct comparison between our work and the different image enhancement methods is not fair. However, in \tableref{mit5k_sota}, we show the results to delineate the need of our approach of style conditioning over conventional CNN based image-to-image mapping to achieve near perfect reconstruction.
}


Next, we train our \ac{INN}-based mapping for the task of inverse tone mapping. While inverse tone-mapping often involves bit-depth expansion and hallucination of over- and under-exposed pixels~\cite{eilertsen2021cheat}, here we focus on the problem of learning global SDR$\rightarrow$HDR color mapping. 
We use the same movie datasets as for the tone-mapping task but swap the source and target frames. The results shown in \figref{results_itm} demonstrate a substantial improvement of 10-12\,dB over HDRNet.

\subsection{Other dimensionality reduction techniques}
\label{sec:pca-vs-autoencoder}

As explained in \secref{problem-formulation}, the color mapping can be expressed as \ac{PCC} (\equationref{style-matrix}) and then the size of the style matrix can be reduced using standard dimensionality reduction methods, such as \ac{PCA} or \ac{VAE}. Here, we compare those standard approaches with our INN. \vspace{0.5em}


\noindent \textbf{Principal component analysis:}
We ran \ac{PCA} on training pairs to reduce the flattened \emph{style} matrix $\mathbf{M}_\textrm{flat}$ into the required number of latent dimensions (2--4). During test time, $\hat{\mathbf{M}}_\textrm{flat}$ is reconstructed from the principal components and used to map the colors. 
We observe from \figref{results_tm} (forward tone mapping) and \figref{results_itm} (inverse tone mapping) that the strict linearity assumption of \ac{PCA} results in poor performance. Qualitative result comparisons are provided in \figref{qualitative_results}. \vspace{0.5em}

\begin{figure*}[t]
    \centering
    \includegraphics[width=\linewidth]{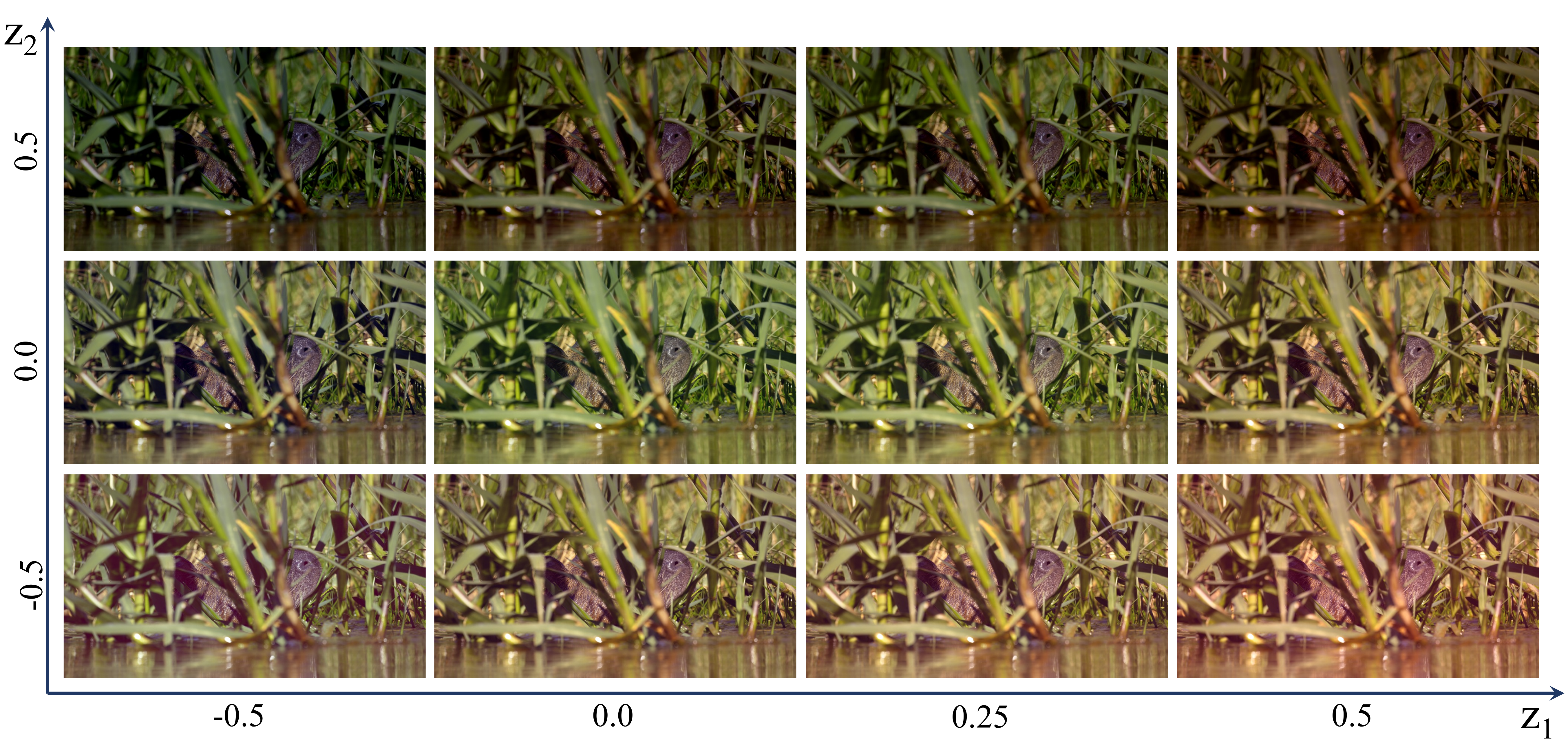}
    \caption{  {Example mapping obtained by manipulating a 2-dimensional \emph{style} vector. Both dimensions control brightness and color temperature. Such style space enables assisted color grading, which mimics the range of styles found in the training image pairs. 
    }
    }
    \label{fig:interpretable-latent}
\end{figure*}

\noindent \textbf{Variational autoencoder:}
Next, we replaced the linear projection with a deep auto-encoder architecture. Since we are interested in an interpretable latent space such as the one depicted in \figref{interpretable-latent}, we opted for a \ac{VAE} where the latent follows a normal distribution~\cite{kingma2013vae}. The input to the \ac{VAE} is the original matrix of polynomial coefficients $\mathbf{M}$. The goals is to train the \ac{VAE} so that the matrix can be predicted from a low-dimensional latent vector. The training loss includes a reconstruction loss between predicted (after matrix multiplication using decoded matrix $\mathbf{\hat{M}}$) and ground truth pixels and the evidence lower-bound. For a fair comparison with our method, we chose a fully connected network with approximately the same number of parameters. We empirically found the best results for a network with 6 hidden layers 
for the encoder and the decoder, with a total of 31K trainable parameters. The network was trained for 500 epochs with an initial learning rate of $5e-4$ with gradual learning rate scheduling. The weights given to the reconstruction and the evidence lower-bound were $\lambda = 1, \delta = 1e-3$, respectively.
Although the \ac{VAE} attains higher quality scores than \ac{PCA} and its results are comparable to HDRNet (which does not use a style vector), \ac{VAE} still performs much worse than our \ac{INN} (see \twofigref{results_tm}{results_itm}).

\section{Applications} \label{sec:applications}

\subsection{Assisted color grading}


\label{subsec:assisted-color-grading}

Color grading is a manual, labor-intensive process that requires a substantial set of skills. Our method can be used to partially automate this process. First, we ask the color artist to manually color grade N scenes, which we use to train our model. Then, we require the color artist to adjust only two or three parameters for the remaining scenes. Such adjustment is much easier than using color grading tools with dozens of different color adjustments. 
The added benefit of using our mapping is that the style is likely to be more consistent across the movie than if a manual color grading tool was used.

Since we do not have access to RAW video frames, typically used for color grading, we demonstrate this application using HDR frames from Blu-ray movies as input and the SDR frames as the color-graded target. Our results, from \secref{results}, have already demonstrated that our mapping can faithfully reproduce the SDR target frames.
Figures~\ref{fig:interpretable-latent} and~\ref{fig:interpretable-latent-islands-lego} show example frames generated by adjusting a 2-dimensional style vector. Each frame comes from a different movie, for which a separate INN was trained. The style space allows for convenient exploration of tone and color adjustments that have been applied to previously color graded frames. The dimensions of the space are easy to interpret: they represent the change of color temperature, contrast and brightness. 
In a supplementary video, we demonstrate a mock interface of the real-time color grading tool.

\subsection{Transmission of SDR and HDR video content}

\begin{figure}[h]
    \centering
    \includegraphics[width=\linewidth]{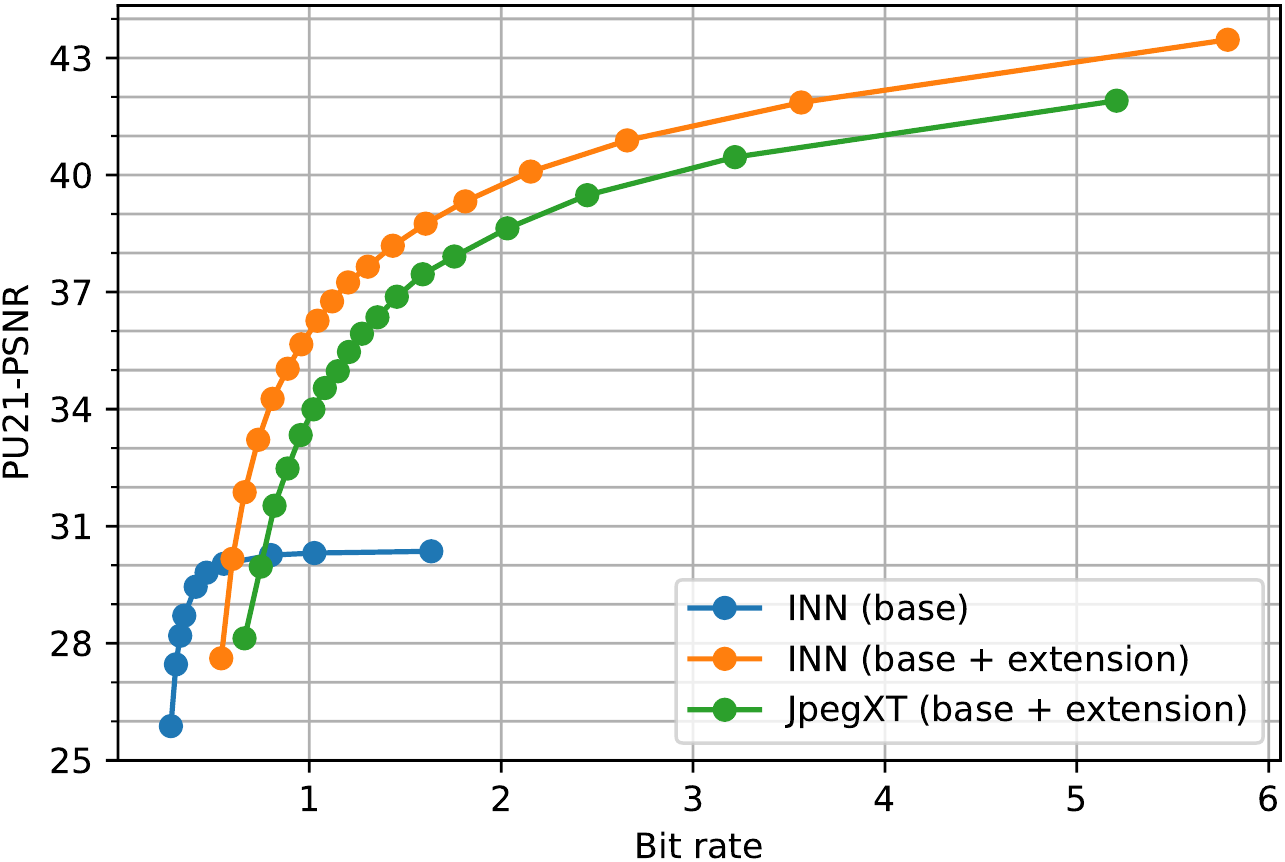}
    \caption{  {Rate-distortion curves, comparing the learned \ac{INN} to JPEG~XT. HDR Image quality (y-axis) is measured by computing PSNR on PU21 encoded images~\cite{mantiuk2021pu21}. 
    }
    }
    \label{fig:compression}
\end{figure}

\begin{figure*}[t]
    \centering
    {\includegraphics[width=0.8\linewidth]{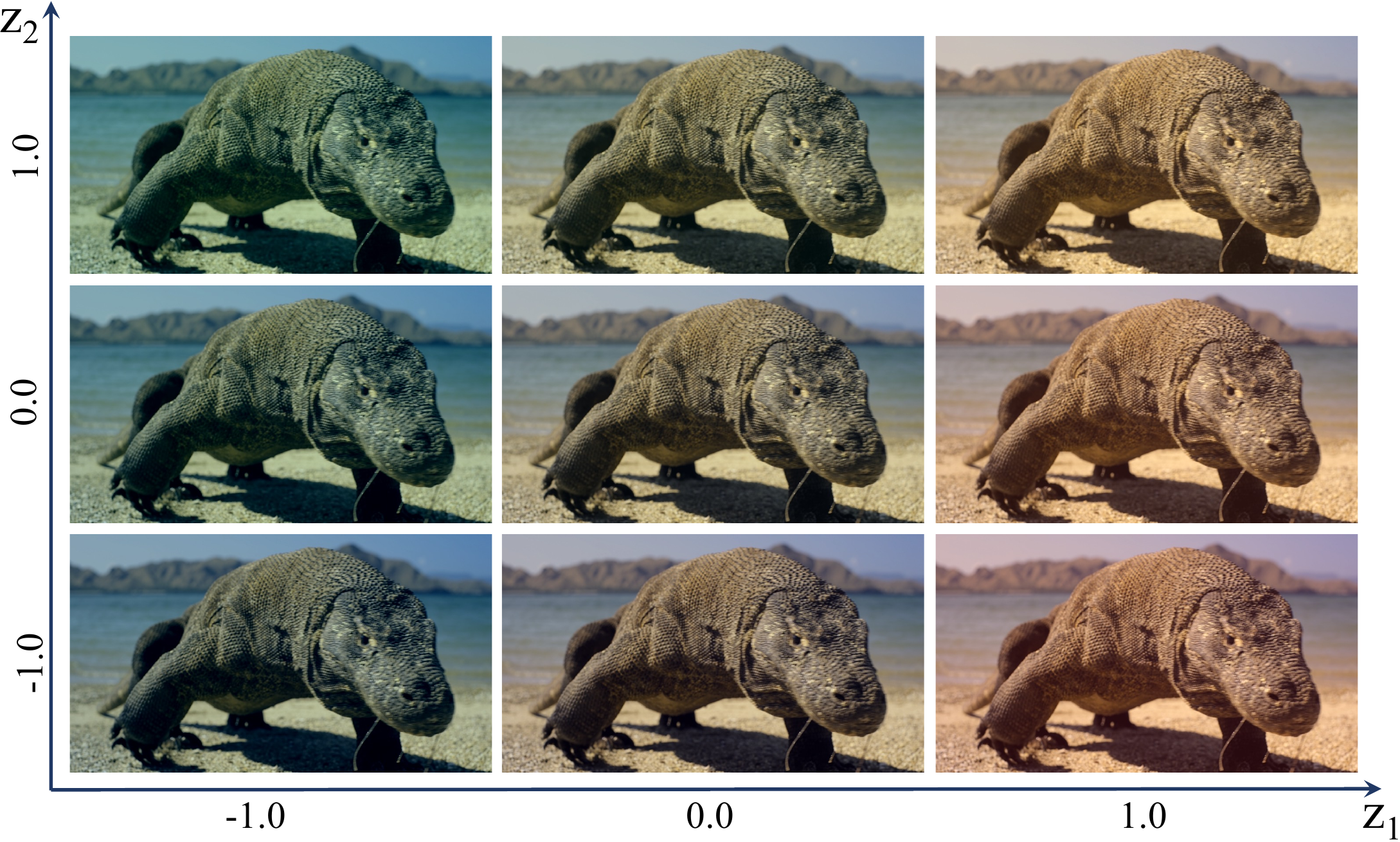} }

\vspace{1em}
    \centering
    {\includegraphics[width=0.8\linewidth]{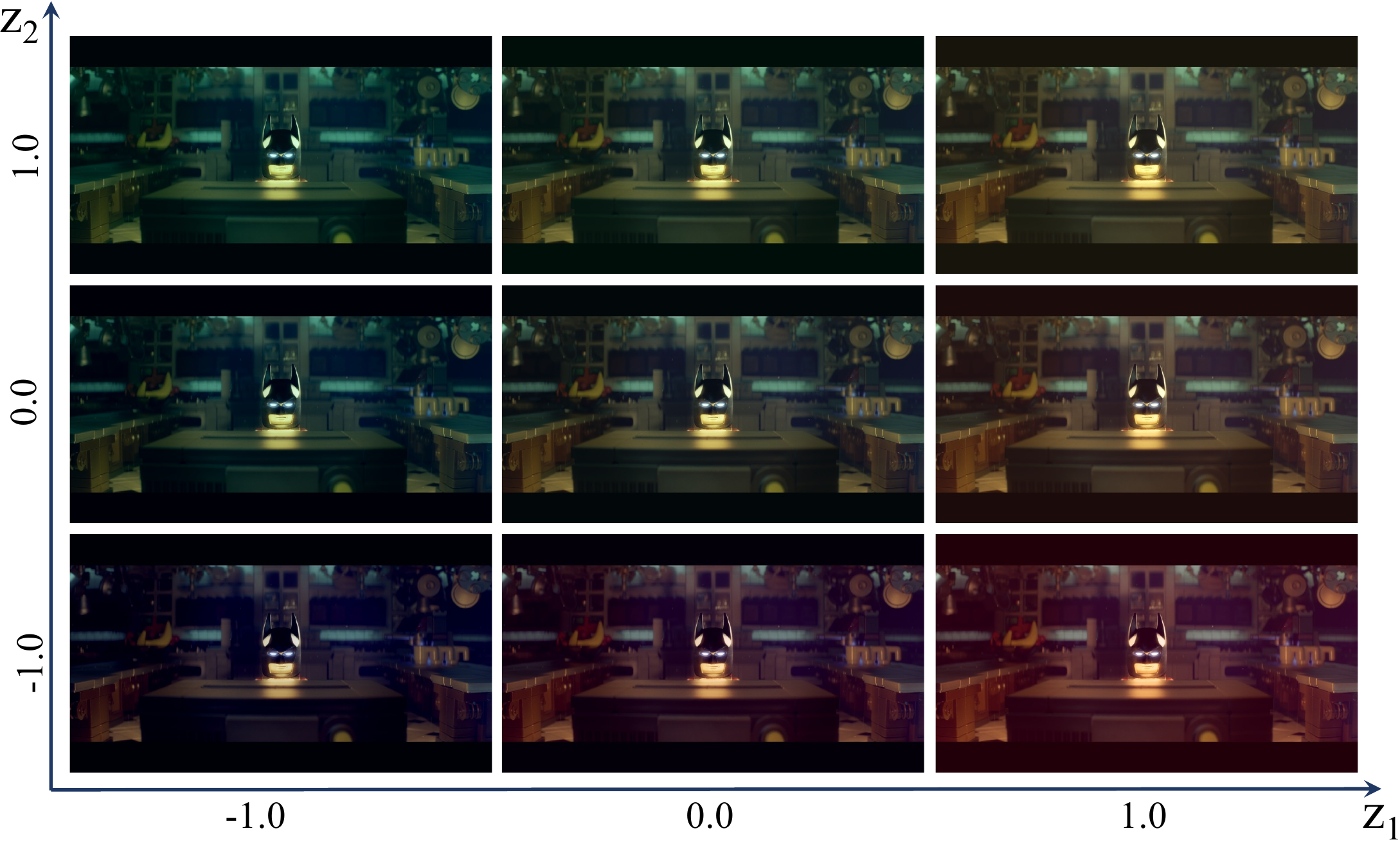} }
    \caption{ {Additional results showing assisted color grading by manipulating a 2-dimensional \emph{style} vector on two datasets - Planet Earth Episode 1 Islands (top) and Lego Batman movie (bottom). Both dimensions control brightness and color temperature. 
    }}
    \label{fig:interpretable-latent-islands-lego}
\end{figure*}


\begin{table*}[t]
\centering
\scriptsize
\begin{minipage}{0.65\textwidth}
\centering

    \begin{tabularx}{\textwidth}{@{}YYYYYYY@{}}
    \toprule
    \rowcolor{black!10} &RGB & PCC-2 & PCC-3 & PCC-4 & PCC-4 +VGG & PCC-4 +Hist  \\
    \midrule
    
    PSNR & 28.98 & 31.52 & 33.02 &  \textbf{41.68} & 30.130 & 35.69 \\
    len($\mathbf{c}_p$) & 3 & 9 & 19 &  34 & 60 & 60 \\

    \bottomrule
    \end{tabularx} 
\end{minipage}
\hfill
\begin{minipage}{0.30\textwidth}
    \begin{tabularx}{\textwidth}{@{}YYYY@{}}
    \toprule
    \rowcolor{black!10} & Dim-2 & Dim-3 & Dim-4 \\ \midrule
    PCA & 17.67 & 18.24 & 18.35 \\
    \ac{VAE} & 20.21 & 28.06 & 31.52\\ 
    \hdashline
    Ours & \textbf{37.85} & \textbf{41.68} & \textbf{41.99}\\

    \bottomrule
    \end{tabularx}
\end{minipage}

\caption{ {Ablation studies over the choice of conditioning vector (left) and over the dimensionality of the latent \emph{style} vector for different methods (right). All results are reported for the ``BBC Planet Earth Jungles" movie in terms of PSNR (dB).}}
\label{tab:ablation_combined}
\end{table*}

The current practice is to encode and distribute SDR and HDR content separately (on Blu-ray or via streaming), which approximately doubles the required storage space. There exist methods for concurrent SDR+HDR image \cite{Artusi2015} and video coding \cite{Mantiuk2006d}, but they require transmitting a substantial amount of additional data. Here we show that our color mapping can substantially reduce, or even eliminate, the need for auxiliary data. 

We compare our inverse tone mapping INN from \secref{results} with the coding used in JPEG~XT (Profile A with open-loop encoding). The frames are encoded individually using either JPEG~XT or a regular JPEG + our learned color mapping. JPEG~XT encodes HDR images by storing a tone-mapped version of an HDR image (base layer), a custom mapping function that predicts the HDR image from the tone-mapped image, and the difference between the predicted and the original HDR image (extension layer). Both base and extension layers are encoded using a standard JPEG codec. We replicate such encoding but replace the custom mapping used in JPEG~XT with our generative color mapping INN. Then, we measure the rate-distortion curves for a test set from the ``Jungles" movie. The rate-distortion curves, shown in \figref{compression}, depict a consistent improvement in performance when using our INN (employing both the base and extension layers). At an extremely low bit rate of 0.5, the INN without an extension layer produces the best quality images (PU-PSNR of over 30~dB).

\subsection{Assisted dynamic range expansion for HDR displays}

The vast majority of video content has been color graded for SDR displays and cannot take advantage of the higher luminance and contrast offered by HDR displays. Here, we show that it is possible to use our mapping function to expand SDR content for HDR displays. It should be noted, however, that our mapping will not be able to reconstruct details in the saturated parts of an image \cite{Eilertsen2017a}. 

This process is identical to color grading, explained in the previous section, except we infer the HDR frames from their SDR counterpart. Similar to forward tone mapping, in this application we require only a small portion of the HDR frames to be manually color graded by color experts.
Results of such assisted dynamic range expansion are included in the appendix.

\section{Ablation study}
\label{sec:ablation}

To achieve the best performance, we conduct an ablation study over the choice of the conditioning vector $\mathbf{c}_p$ for training the \ac{INN}. Furthermore, we provide a study over the dimensionality of the \emph{style} vector.

\subsection{Conditioning vector}
\label{sec:ablation-conditioning}

First, we study the effect of using different degrees of polynomials in PCC as our per-pixel conditioning vector. We conducted an ablation study over the 1$^\textnormal{st}$ (RGB), 2$^\textnormal{nd}$, 3$^\textnormal{rd}$ and 4$^\textnormal{th}$ degree polynomial expansion. Second, we study the effect of using image statistics as the conditioning vector in addition to the 4$^\textnormal{th}$ degree polynomial. We train our INN architecture in such that each pixel-wise conditioning vector (PCC) is concatenated with high level features extracted from the entire image using an additional feed-forward network $H$. For this task, we use a pre-trained VGG network, similar to \cite{ardizzone2020conditional,denker2021conditional} as the feed-forward network. The final conditioning vector for a given pixel is given as  $\mathbf{c}_p = [ \mathcal{C}(\mathbf{y}_p), H(\mathbf{y})]$, where the weights of $H(\cdot)$ are simultaneously being updated alongside the weights of the \ac{INN}. Finally, we train our INN using the 4$^\textnormal{th}$ degree polynomial concatenated with the histogram of the luma channel as the conditioning vector. \tableref{ablation_combined} (left) shows that the \ac{INN} performs best when no additional image statistics are added to the 4\textsuperscript{th}-degree PCC conditioning vector. Note that different conditioning vectors have different lengths, as shown in the second row of~\tableref{ablation_combined} (left).

\subsection{Dimensionality of the style vector}
\label{sec:ablation-dimensions}

Next, we investigate the impact of changing the dimensionality of the latent \emph{style} vector. Mapping the style of a target domain image to a low dimensional latent representation allows easy user-interactive image manipulation. To this end, we provide additional network architectures in~\secref{dim-style-code} to train our INN for 2 and 4 latent style encodings. In \tableref{ablation_combined} (right) we further provide a comparison of similar dimensions of the latent representations for \ac{PCA} and \ac{VAE}.

\section{Conclusions}

This work highlights the importance of modeling style when learning global image transforms, such as those between differently color-graded SDR-HDR images. We conclusively show that extracting a \emph{style} vector from a target image considerably improves the reconstruction quality. This is due to the existence of an infinite number of equally plausible solutions, each representing a unique color artist's choice. Our proposed conditional \ac{INN} effectively models this one-to-many mapping by extracting and encoding this artistic choice from examples of image pairs into a low-dimensional \emph{style} vector. We show that our method
significantly outperforms state-of-the-art deep architectures that ignore style, as well as alternate dimensionality reduction methods that incorporate latent style but cannot encode it efficiently. Moreover, our invertible framework enables interactive style manipulation by adjusting the low-dimensional latent vector. The main focus of our work is color mapping for video content, which is global in nature (the same for all pixels in an image, $\mathcal{M}: \mathcal{R}^3 \rightarrow \mathcal{R}^3$). However, our method can be easily extended to a manual local mapping by splitting frames into a number of tiles, as done in \cite{Eilertsen2015}.

\section*{Acknowledgements}
\noindent This project has received funding from the European Research Council (ERC) under the European Union’s Horizon 2020 research and innovation programme (grant agreement N$^\circ$ 725253–EyeCode).

\clearpage

\section*{\Large Appendix}
\setcounter{section}{0}

In this appendix, we further report quantitative results for both forward and inverse tone-mapping on different datasets using additional metrics (\secref{extra-results}). We also investigate whether a single INN can effectively capture all the expert styles from the \FiveK dataset (\secref{mit5k-all}). Finally in~\secref{bi-directional-training}, we show improvement in performance due to the proposed bi-directional training with \ac{NLL} $\mathcal{L}_\textrm{NLL}$ and the reconstruction loss $\mathcal{L}_\textrm{rec}$.





\section{Additional results}
\label{sec:extra-results}
\subsection{Forward tone mapping}

\begin{figure*}[t!]
    \centering
    {\includegraphics[width=0.99\linewidth]{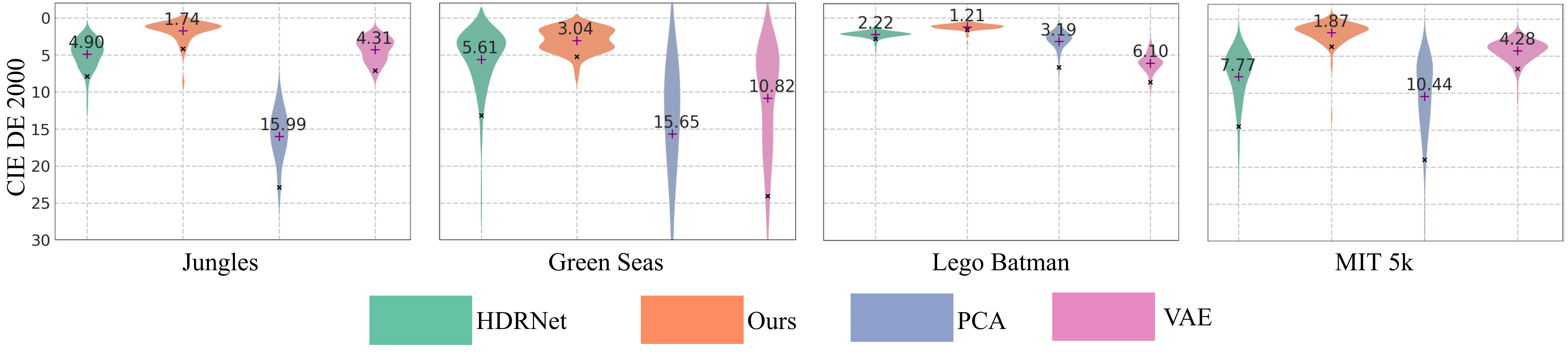} }
    \caption{ \small{Comparison of results on different datasets for the task of forward tone mapping for \CIEDE metric. Our method achieves a substantial improvement in performance compared to other dimensionality reduction methods across datasets. The purple `$+$' in the plots show the mean and black `$\times$' show the lowest 5$^{th}$ percentiles. Note that the y-axis for \CIEDE metric has been reversed. }}
    \label{fig:sup-ftm}
\vspace{1em}
    \centering
    {\includegraphics[width=0.99\linewidth]{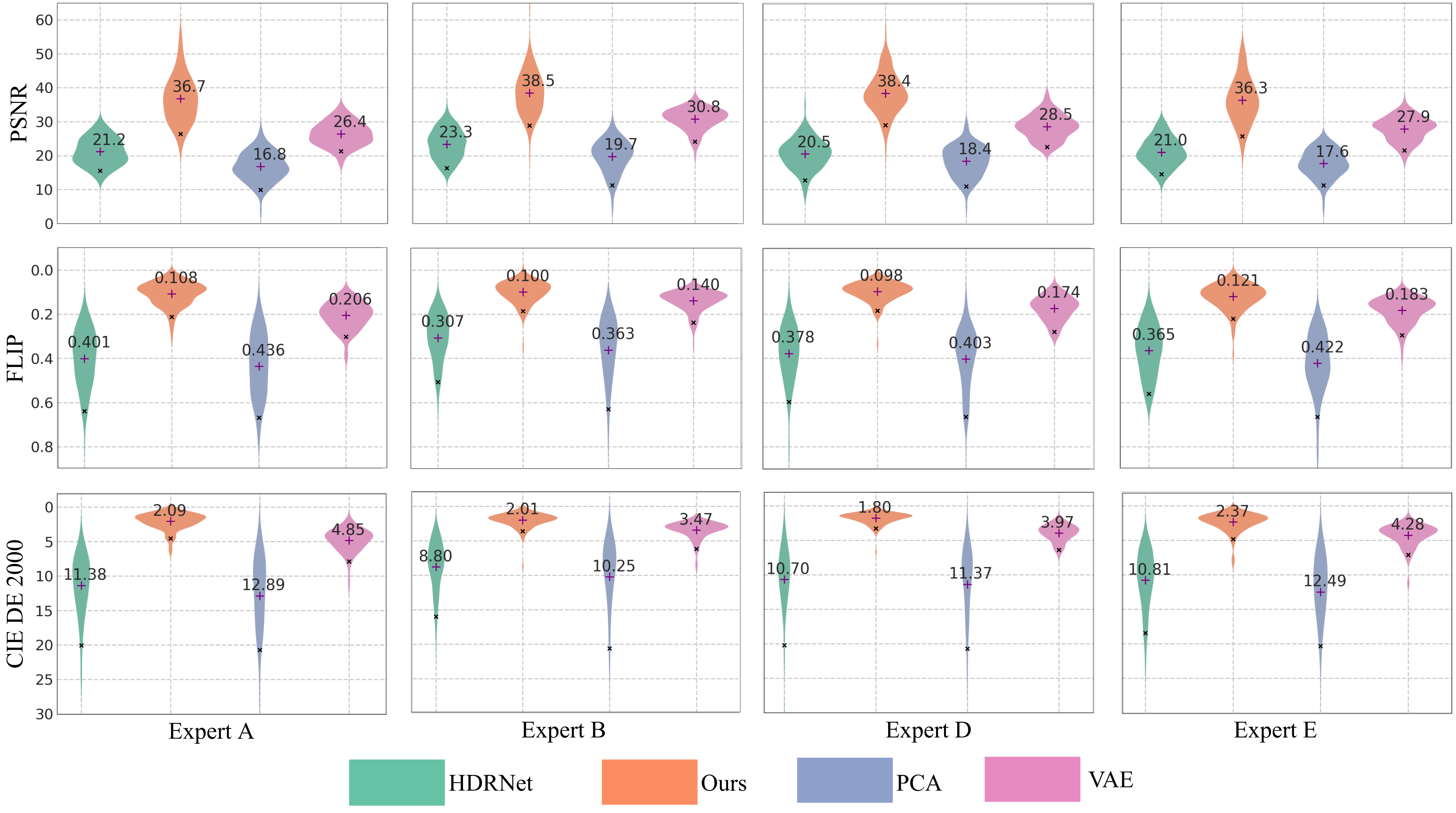} }
    \caption{ \small{Comparison of our results with other methods on the remaining 4 experts from the \FiveK dataset. Note that our method achieves a substantial improvement in performance compared to other dimensionality reduction methods across experts. The purple `$+$' in the plots show the mean and black `$\times$' show the lowest 5$^{th}$ percentiles. Note that the y-axis for FLIP and \CIEDE metric have been reversed. }}
    \label{fig:sup-mit5k-individual}
\end{figure*}

The violin plots in \figref{sup-ftm} compare our INN with HDRNet, PCA and VAE using the \CIEDE~\cite{sharma2005ciede2000} metric. Similar to Fig. 6 in the main document, there is a substantial improvement in reconstruction quality due to correctly extracting and utilizing style.

For the \FiveK dataset, we provide results for all the experts in~\figref{sup-mit5k-individual}. These are consistent with images retouched by expert C reported in the other figures. Here, a separate network is used for each expert to better compare with existing works.



\subsection{Inverse tone mapping}
\begin{figure*}[t!]
    \centering
    {\includegraphics[width=0.99\linewidth]{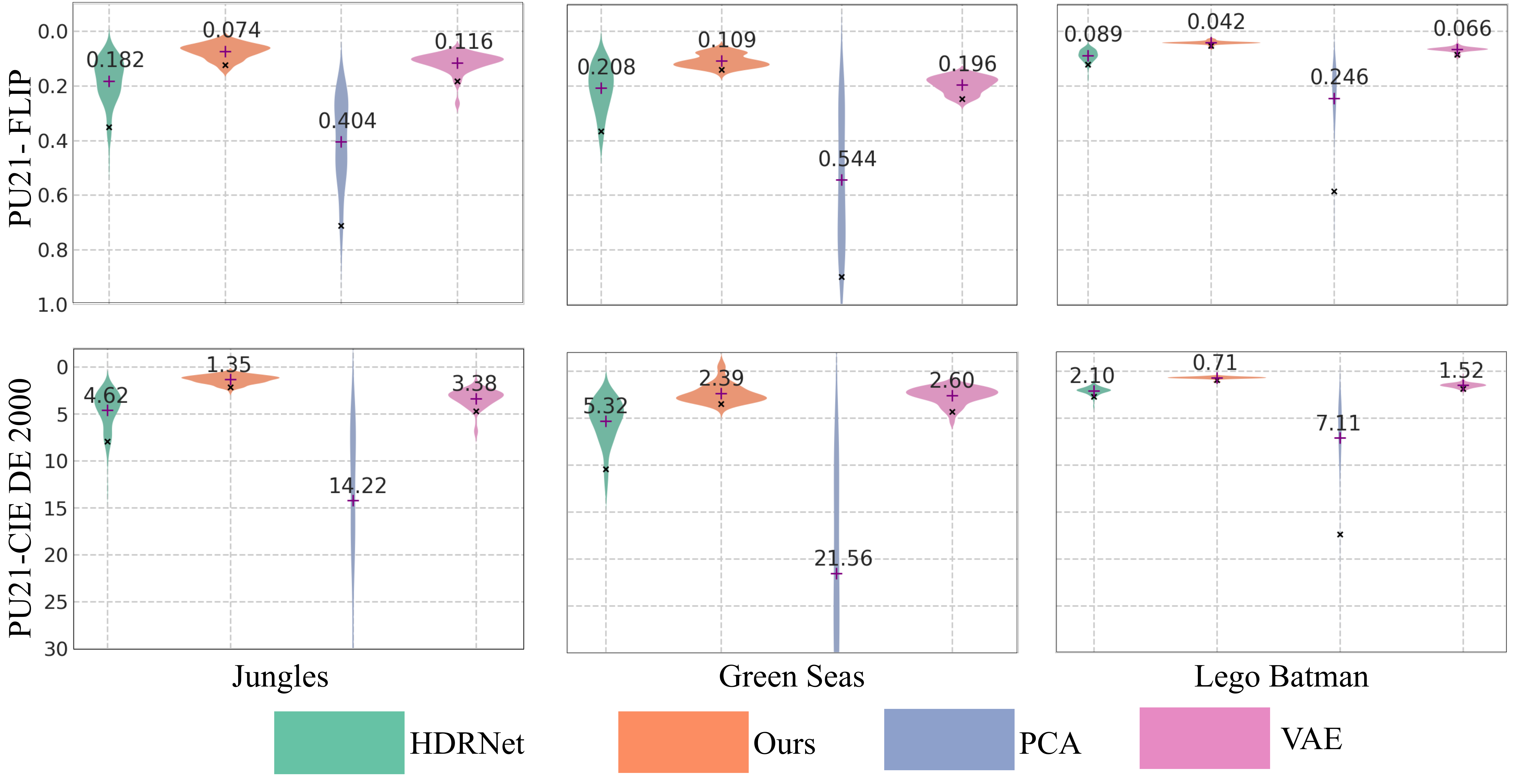} }
    \caption{ \small{Here we show the results for the task of inverse tone mapping on two other metrics: FLIP and \CIEDE. To adapt the metrics to HDR images, we use perceptually uniform PU21 transform~\cite{mantiuk2021pu21}. The purple `$+$' in the plots show the mean and black `$\times$' show the lowest 5$^{th}$ percentiles. Note that the y-axis for FLIP and \CIEDE metric have been reversed. }}
    \label{fig:sub-inverse_tone_mapping}
    
\end{figure*}

For the task of inverse tone mapping,~\figref{sub-inverse_tone_mapping} shows similar violin plots for 2 more metrics: FLIP and \CIEDE. Before running the \ac{SDR} metrics, we encode the reconstructed and ground truth HDR images with PU21 encoding.

\subsection{Image content vs. style conditioning}
\label{sec:mit5k-all}

\begin{figure*}[h]
    \centering
    {\includegraphics[width=0.99\linewidth]{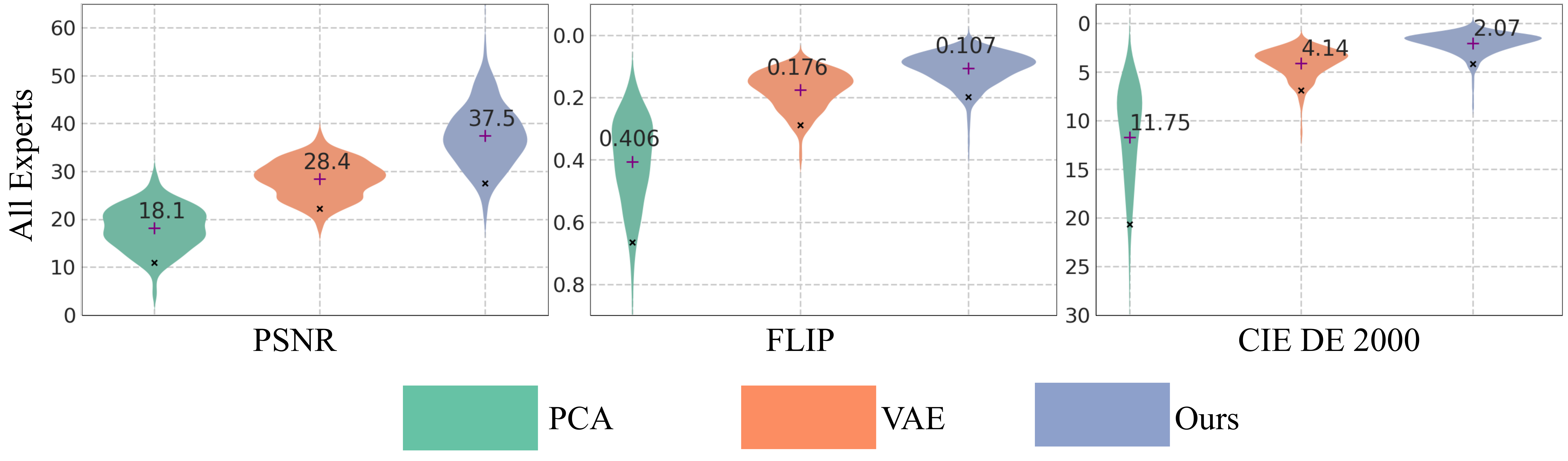} }
    \caption{ \small{Here we show a comparison of our method with PCA and VAE when a single model is trained for all experts from the \FiveK dataset. We report the results in terms of PSNR, FLIP and \CIEDE~metric. Note that the y-axis for FLIP and \CIEDE metric have been reversed. The test set in this case contains 5000 images, 1000 for each expert. Additional inference results for each expert individually are reported in \tableref{experts_individually}.}}
    \label{fig:sup-mit5k-combined}
\end{figure*}
\begin{figure*}[t!]
    \centering
    {\includegraphics[width=0.99\linewidth]{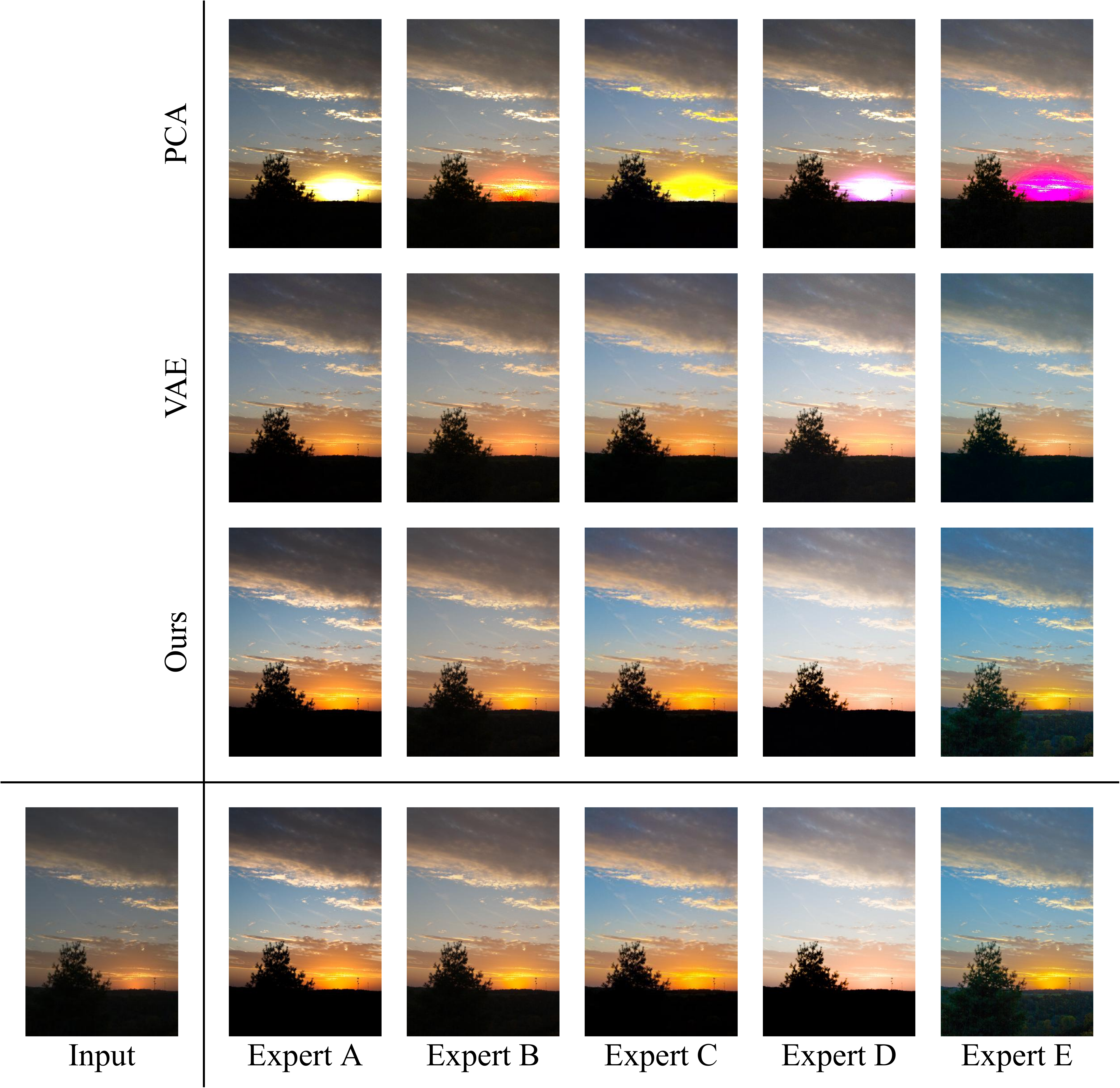} }
    \caption{\small{Qualitative comparison of methods that simultaneously capture the different styles of all experts of the \FiveK dataset. The bottom row shows the reference images. Quantitative results over all test images are plotted in~\figref{sup-mit5k-combined}. Note that HDRNet is unsuitable for this task since a single network cannot learn different styles corresponding to the experts.}}
    \label{fig:sup-grid_1_mit5k}
\end{figure*}
\begin{figure*}
    \centering
    {\includegraphics[width=0.99\linewidth]{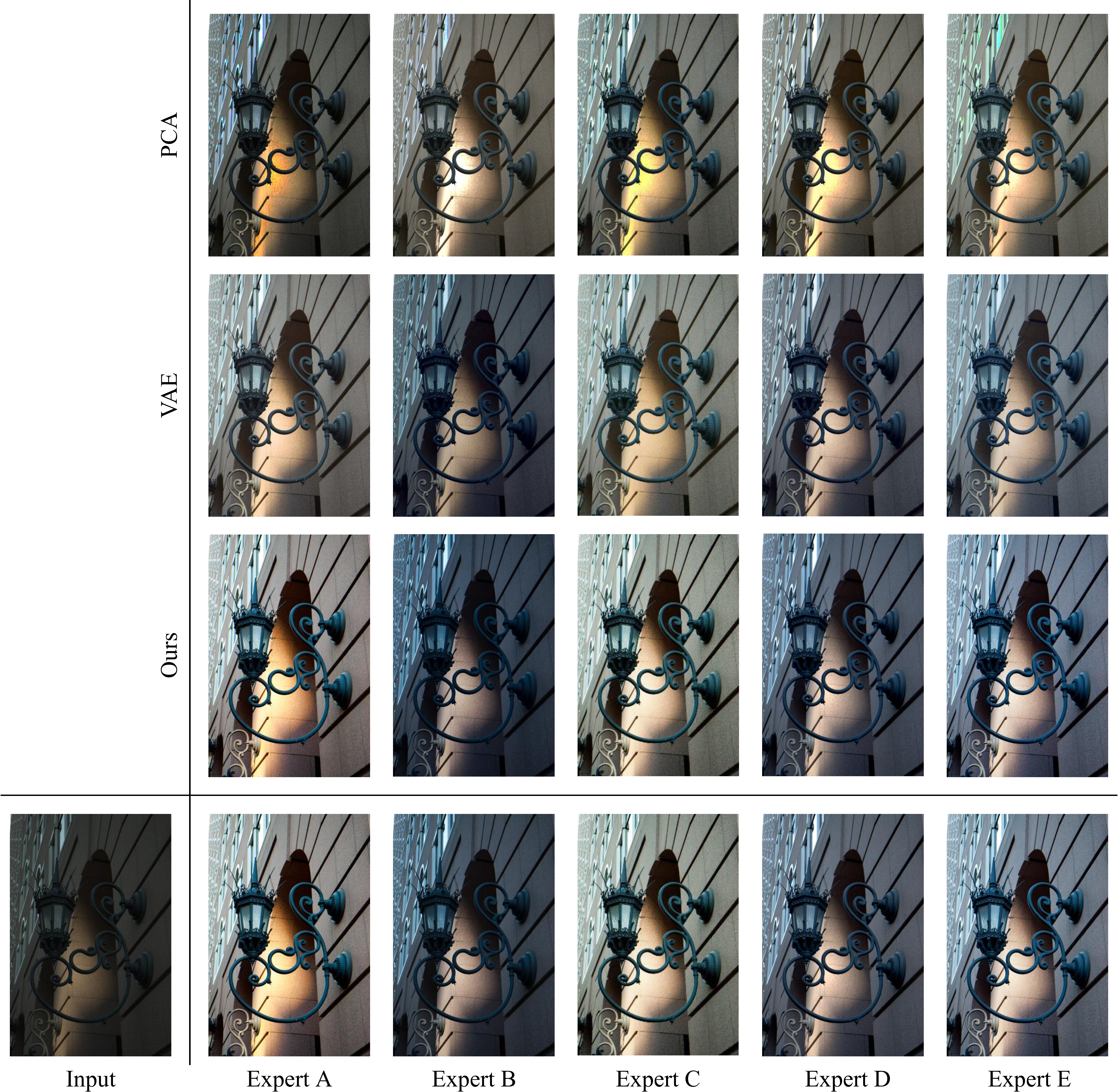} }
    \caption{ \small{Here we show qualitative comparison of methods that simultaneously capture the different styles of all experts of the \FiveK dataset for a different image. The bottom row shows the reference images. }}
    \label{fig:sup-grid_2_mit5k}
\end{figure*}

Additionally, we trained a single INN for all 5 experts of \FiveK, something that can not be done with HDRNet because of the lack of conditioning on style. Since we learn a one-to-many mapping, the same network produces outputs in different styles by utilising different latent vectors. \figref{sup-mit5k-combined} shows that our single INN successfully captures the style of all the experts. We further report the quantitative comparison of our single trained INN on the individual test set for each expert in \tableref{experts_individually}. Refer to \figref{sup-grid_1_mit5k} and \figref{sup-grid_2_mit5k} for qualitative comparisons on selected scenes. The deep-learning methods, like HDRNet, that model expert retouching with one-to-one mappings are unsuitable for this task since a single network cannot learn different styles corresponding to the experts. We see that our INN performs much better than PCA and VAE and produces artifact-free images that better match the required style.

\section{Bi-directional Training}
\label{sec:bi-directional-training}

When trained with the \ac{NLL} loss $\mathcal{L}_\textrm{NLL}$, our INN learns to conditionally map a target pixel to a latent vector. However, when presented with an entire frame, there is no easy way to extract a single low-dimensional vector, that captures the style of the mapped frames. Recall that we augment \ac{MLE} training by forcing the per-pixel representations of the same frame to lie closer in the latent space.

In \tableref{bi_directional_training}, we show the effect of our proposed bi-directional training, by addition of the reconstruction loss $\mathcal{L}_\textrm{rec}$ alongside $\mathcal{L}_\textrm{NLL}$. The NLL loss makes the style vectors resemble a predetermined latent distribution (the standard normal in our experiments), while the reconstruction loss ensures that pixels from the same frame have similar style vectors.

\begin{table}[h]

\centering

\caption{\small{Ablation study on the effect of bi-directional training for the task of forward tone-mapping for the ``BBC Planet Earth II Episode 3 - Jungles" dataset.}} 
\vspace{1em}
\label{tab:bi_directional_training}
\begin{tabularx}{0.5\textwidth}{|Y|Y|Y|}

    \hline \rowcolor{black!10}
    & $\mathcal{L}_\textrm{NLL}$ & $\mathcal{L}_\textrm{NLL} + \mathcal{L}_\textrm{rec}$ \\
    \hline \hline
    PSNR $\uparrow$ & 36.80 & \textbf{41.68}  \\
    FLIP $\downarrow$ &0.108  &  \textbf{0.070}\\
    \hline
\end{tabularx}

\end{table}

\begin{table*}[h]

\centering

\caption{\small{Additional quantitative results for each expert from the \FiveK dataset for a single model trained on all experts. Results are reported in terms of PSNR, FLIP and \CIEDE~metric. Note that HDRNet is unsuitable for this task since a single network cannot learn different styles corresponding to the experts.}} 
\vspace{1em}
\label{tab:experts_individually}
\begin{tabularx}{0.995\textwidth}{|Y|Y|Y|Y|Y|Y|}

    \hline \rowcolor{black!10}
    & Expert A & Expert B & Expert C & Expert D & Expert E\\
    \hline \hline
    PSNR $\uparrow$ & 36.4 & 38.1 & 38.9 & 38.2& 36.0\\
    FLIP $\downarrow$ & 0.116 & 0.104 & 0.099& 0.100 &0.124  \\
    CIE DE $\downarrow$ & 2.11 & 2.02 & 1.91& 1.93&2.38  \\
    \hline
\end{tabularx}

\end{table*}

\bibliographystyle{ACM-Reference-Format}

\clearpage
\clearpage
\bibliography{references}










\end{document}